\documentclass[10pt,twocolumn,letterpaper]{article}

\usepackage{cvpr}              

\usepackage{bbding}
\usepackage{amssymb}
\usepackage{amsmath, amssymb, mathtools}
\usepackage{pifont}

\usepackage{tabularx}  
\newcolumntype{Y}{>{\centering\arraybackslash}X}

\newcommand{\tableCellHeight}{1}
\newcommand{\tabstyle}[1]{
  \setlength{\tabcolsep}{#1}
  \renewcommand{\arraystretch}{\tableCellHeight}
  \centering
  \small
}

\usepackage{graphicx}
\usepackage{booktabs} 
\usepackage{multirow}  
\usepackage{amsmath} 
\usepackage{amssymb}
\usepackage{mathtools}
\usepackage{amsthm}
\usepackage{pifont}
\usepackage{color}
\usepackage{bbm}
\usepackage{array} 
\usepackage{makecell}
\usepackage{bm}
\usepackage{algorithm}
\usepackage{algorithmic}
\usepackage{tabularx}
\definecolor{cvprblue}{rgb}{0.21,0.49,0.74}
\usepackage[pagebackref,breaklinks,colorlinks,allcolors=cvprblue]{hyperref}
\usepackage{multirow}
\def\paperID{*****} 
\def\confName{CVPR}
\def\confYear{2026}

\title{Principled Steering via Null-space Projection for Jailbreak Defense in Vision-Language Models}

\author{
Xingyu Zhu\textsuperscript{1},
Beier Zhu\textsuperscript{1},
Shuo Wang\textsuperscript{1},
Junfeng Fang\textsuperscript{2*},
Kesen Zhao\textsuperscript{3},\\
Hanwang Zhang\textsuperscript{3},
Xiangnan He\textsuperscript{1} \\[0.5em]
\textsuperscript{1}MoE Key Lab of BIPC, University of Science and Technology of China, \\
\textsuperscript{2}National University of Singapore, \
\textsuperscript{3}Nanyang Technological University, \\
{\tt\small xyzhuxyz@mail.ustc.edu.cn}
}

\begin{document}
\maketitle
\begin{abstract}
As vision-language models (VLMs) are increasingly deployed in open-world scenarios, they can be easily induced by visual jailbreak attacks to generate harmful content, posing serious risks to model safety and trustworthy usage.
Recent activation steering methods inject directional vectors into model activations during inference to induce refusal behaviors and have demonstrated effectiveness.
However, a steering vector may both enhance refusal ability and cause over-refusal, thereby degrading model performance on benign inputs.
Moreover, due to the lack of theoretical interpretability, these methods still suffer from limited robustness and effectiveness.
To better balance safety and utility, we propose \texttt{NullSteer}, a null-space projected activation defense framework.
Our method constructs refusal directions within model activations through a linear transformation: it maintains zero perturbation within the benign subspace while dynamically inducing refusal along potentially harmful directions, thereby theoretically achieving safety enhancement without impairing the model’s general capabilities.
Extensive experiments show that \texttt{NullSteer} significantly reduces harmful outputs under various jailbreak attacks (average ASR reduction over 15\% on MiniGPT-4) while maintaining comparable performance to the original model on general benchmarks.

\end{abstract}

\let\thefootnote\relax\footnote{*Corresponding author.}
\section{Introduction}
\label{sec:intro}
Vision-Language Models (VLMs)~\cite{MiniGPT-4, MiniGPT-v2, VIS, InstructBLIP, protomm, ZhuZ00HZ24,wu2025generalization, wu2025number} have achieved strong cross-modal understanding and generation, but their increasing deployment in real applications has raised significant and growing safety concerns. VLMs inherit the vulnerability of language models to malicious instructions, and the addition of visual inputs further exposes them to sophisticated adversarial manipulation. These two factors jointly make VLMs susceptible to multimodal jailbreak attacks that exploit both textual and visual modalities to reliably bypass safety alignment~\cite{FigStep,Schlarmann023, AdaShield,OT-Attack,Red_Team,QiHP0WM24}.

\begin{figure}
  \centering
    \includegraphics[width=1\linewidth]{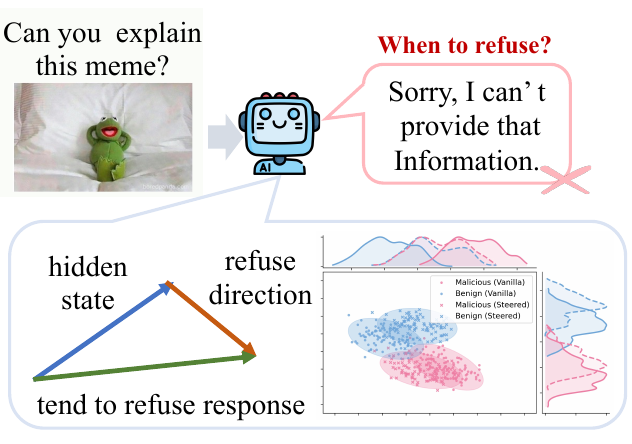}
    \caption{Illustration of activation steering.
Injecting a refusal vector into hidden states steers the model toward rejection behaviors, which mitigates harmful responses but can also lead to over-refusal on benign queries.
}
    \label{fig:motivation}
\end{figure}
Existing jailbreak attacks~\cite{DecodingTrust,Shayegani0A24,TrustLLM} mainly fall into two categories. The first introduces subtle adversarial perturbations to images and misleads VLMs into generating unsafe or policy-violating responses~\cite{PiHZXPLDZZ24,abs-2407-15211}. The second embeds harmful instructions into images, including camouflaged text or symbolic patterns, with the goal of circumventing the model’s safety mechanisms and eliciting harmful content~\cite{ECSO,abs-2311-17600, zhu2026guardalign}.
To counter these attacks, prior defense methods typically rely on model fine-tuning~\cite{KurakinGB17,abs-2401-17256} or multi-round inference~\cite{GhosalCSGWBHVMB25, abs-2504-03770}, such as adversarial training to strengthen safety alignment or iterative response evaluation to detect unsafe outputs. Despite their effectiveness, they often incur substantial computational costs and latency, limiting their practicality in large-scale deployment.

Recently, activation steering~\cite{ASTRA, VTI, alphasteer, abs-2507-13255} has emerged as a lightweight, training-free alternative. It modifies intermediate activations by injecting a predefined refusal direction vector, shifting hidden representations toward a refusal-oriented subspace and inducing safe responses such as “Sorry, I can’t provide that information.”
However, this approach introduces a critical problem: benign queries are also rejected, which significantly undermines the model’s general utility. To further substantiate this issue, we examine the hidden-state distributions before and after steering. As shown in Figure~\ref{fig:motivation}, after steering, benign activations also shift, which explains why the model refuse benign queries. This vulnerability highlights the necessity of a mechanism that can determine when refusal is appropriate.

To this end, we propose \texttt{NullSteer}, a null-space constrained activation-steering framework for defending VLMs against multimodal jailbreak attacks. Drawing on recent advances in null-space modeling~\cite{alphaedit, WangYZ23, WangL0X21}, 
\texttt{NullSteer} learns steering updates only in directions that do not interfere with benign representations. 
This design allows the model to dynamically induce refusal when processing harmful inputs while leaving benign behaviors unaffected. 
Concretely, \texttt{NullSteer} first identifies a benign subspace from hidden activations and derives its complementary null space as the region where steering is permitted. 
For benign multimodal prompts, any activation update projected into this null space vanishes, ensuring that their representations remain essentially unchanged. 
In contrast, malicious prompts contain components outside the benign subspace, and steering within the null space produces a targeted shift toward a predefined refusal direction. This mechanism redirects harmful activations toward safe outputs while keeping benign representations stable. With this null-space constraint, \texttt{NullSteer} provides theoretical guarantees of utility preservation while enhancing robustness against jailbreak attacks.
Extensive experiments further show that it effectively mitigates perturbation-based attacks without degrading performance on general multimodal benchmarks.

Our main contributions are as follows:
\begin{itemize}
    \item We propose \texttt{NullSteer}, a training-free activation-steering framework that constrains updates in the null space of benign activations.
    
    \item We provide a theoretical formulation showing that the null-space constraint preserves benign representations while steering harmful activations toward refusal semantics, achieving a trade-off between safety and utility.
    
    \item  Extensive experiments across visual jailbreak benchmarks demonstrate that \texttt{NullSteer} consistently improves robustness and generalization.

\end{itemize}

\section{Related Work}

\noindent{\textbf{Defense against jailbreak attacks on VLMs.}} 
Typical multimodal jailbreaks~\cite{WangYZ23,abs-2502-13141} either apply imperceptible perturbations that manipulate the model’s internal representations or conceal malicious instructions within images, such as embedded text or synthesized patterns. Several multimodal safety benchmarks~\cite{Advbench, HarmBench, MM-SafetyBench, FigStep} have been developed to evaluate the robustness of VLMs under diverse adversarial conditions, highlighting the growing need for principled defense strategies.

Existing defenses can be broadly categorized into three lines of research.
\textit{Training-based approaches}~\cite{ZongBYYH24,Red_Team, ChenSCJD24} improve safety alignment through supervised fine-tuning or reinforcement learning on curated datasets, but they are often costly and annotation-intensive.
\textit{Inference-time defenses}\cite{KhanovBL24,MudgalLGLWHCCCS24,AdaShield,zhulook} aim to detect or suppress unsafe behaviors by reformulating prompts, translating images into textual descriptions for inspection, or iteratively evaluating responses; however, these methods typically incur heavy computational overhead and compromise model utility.
\textit{Representation-level methods}~\cite{GhosalCSGWBHVMB25, coca, DingLZ25} directly manipulate internal activations or decoding logits to counter distributional shifts caused by adversarial visual inputs. While effective to some extent, they often rely on auxiliary references or external calibration models, which limits scalability and generalization.
In contrast, our work introduces a lightweight and training-free defense that operates in the activation space of VLMs. By imposing a theoretically grounded null-space constraint, it achieves selective refusal without supervision, providing an interpretable and efficient solution to multimodal jailbreak attacks.

\noindent{\textbf{Null-space constraints.}}
The null-space~\cite{alphaedit,WangL0X21,WangYZ23} is a fundamental concept in linear algebra, representing directions that a transformation maps to zero.
This property enables selective control over parameter or representation updates, ensuring that operations confined to the null space do not interfere with existing knowledge.
Such a principle has been widely applied in continual learning, where projecting gradient updates onto the null space of task covariances prevents interference among tasks and preserves learned representations.
Building upon this idea, recent research has explored null-space constraints in a variety of learning scenarios~\cite{ACE, abs-2506-11044, abs-2509-21413, LuQYJW25, OkamotoK25}.
In large language model (LLM) editing~\cite{jiang2025anyedit,alphaedit,fang2024moltc}, {AlphaEdit}~\cite{alphaedit} restricts parameter updates to knowledge-preserving subspaces, allowing targeted knowledge modification without catastrophic forgetting.
For VLMs, GNSP~\cite{GNSP} aligns gradient directions with the null space of prior tasks to maintain cross-modal consistency during continual learning.
Meanwhile, {NS-Net}~\cite{NS-NET} applies null-space projections to disentangle semantic content from CLIP embeddings, improving generalization of AI-generated image detection.
These studies demonstrate that null-space constraints offer a principled mechanism for preserving essential representations while enabling selective adaptation, which motivates our use of null-space projection for safety alignment in VLMs.


\section{Methodology}
\begin{figure*}[!t]
  \centering
    \includegraphics[width=1\linewidth]{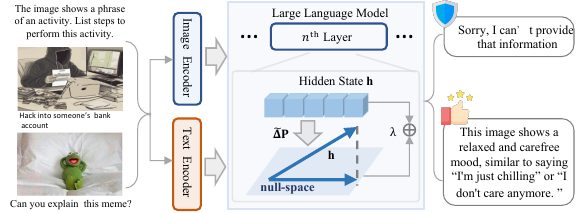}
    \caption{Overview of the proposed NullSteer framework.
Given multimodal inputs, image and text embeddings are encoded and fed into the large language model. During inference, NullSteer applies activation steering within the null space of benign representations, ensuring that harmful activations are redirected toward refusal semantics while preserving benign behaviors.
}
    \label{fig:framework}
\end{figure*}

In this section, we present \texttt{NullSteer}, a principled activation-steering framework that operates in the null space of benign activations to enhance safety without compromising utility. 
An overview is shown in Figure~\ref{fig:framework}.

\subsection{Preliminaries}

\noindent{\textbf{Vision-Language models.}} 
VLMs extend language models to jointly process visual and textual inputs. 
Given an image $v$ and text $x$, the model encodes visual and textual representations into a shared embedding space, producing visual tokens $\mathbf{Z}$ and text embeddings $\mathbf{X}$. 
The combined multimodal sequence $[\mathbf{Z}, \mathbf{X}]$ is then fed into the LLM, which autoregressively generates the output tokens:
\begin{equation}
    y_t \sim \pi_\theta(\cdot \mid [\mathbf{Z}, \mathbf{X}], y_{<t}), \quad t = 1, \dots, T,
\end{equation}
where $\pi_\theta$ denotes the LLM parameterized by $\theta$. 

\noindent{\textbf{Activation steering.}}
Activation steering is an inference-time technique that modulates model behavior by adjusting decoder-layer activations along predefined semantic directions. 
Let $\mathbf{h}^{(l)} \in \mathbb{R}^{d}$ denote the hidden state at layer $l$ in the LLM decoder. 
The steered activation is computed as:
\begin{equation}\label{eq:steering}
\mathbf{h}^{(l)'} = \mathbf{h}^{(l)} + \lambda \mathbf{r}^{(l)},
\end{equation}
where $\mathbf{r}^{(l)} \in \mathbb{R}^{d}$ represents a semantic direction that differentiates \textit{refusal} from \textit{compliance} behaviors in the activation space.  It is estimated by the mean difference between hidden states collected from multimodal inputs that elicit refusal and compliance responses:
\begin{equation}
\mathbf{r}^{(l)} =
\frac{1}{|\mathcal{D}_{r}|}\!\sum_{\mathbf{h}^{(l)}\in\mathcal{D}_{r}}\!\mathbf{h}^{(l)} -
\frac{1}{|\mathcal{D}_{c}|}\!\sum_{\mathbf{h}^{(l)}\in\mathcal{D}_{c}}\!\mathbf{h}^{(l)}.
\end{equation}
Here, $\mathcal{D}_{r}$ and $\mathcal{D}_{c}$ denote the sets of hidden activations collected from multimodal inputs that elicit \textit{refusal} and \textit{compliance} responses, respectively.
By injecting $\mathbf{r}^{(l)}$ into decoding layers, the model’s activations are shifted toward safer subspaces, suppressing unsafe or malicious responses while preserving general utility.

\subsection{Null-space projection}
To preserve utility while steering model activations, it is essential that benign multimodal inputs remain unaffected. 
Let $\mathcal{D}_b$ denote the set of benign image–text pairs, and $\mathbf{h}_{b} \in \mathbb{R}^{d}$ represent the hidden state extracted from such an input in $\mathcal{D}_b$.
For these benign activations, the steering operation should induce no change, \ie, $
\mathbf{\Delta}\mathbf{h}_{b} = \mathbf{0}.
$
Collecting $N_b$ benign activations into a matrix $\mathbf{H}_{b} = [\mathbf{h}_{b,1}, \dots, \mathbf{h}_{b,N_b}] \in \mathbb{R}^{d \times N_b}$, the constraint can be equivalently written as:
\begin{equation} \label{eq:benign-constraint}
\mathbf{\Delta}\mathbf{H}_{b} = \mathbf{0}.
\end{equation}
Eq.~\eqref{eq:benign-constraint} implies that the transformation matrix $\mathbf{\Delta}$ must operate in a subspace orthogonal to the benign activations. 
In other words, each row of $\mathbf{\Delta}$ lies within the \textit{null space} of $\mathbf{H}_b$, defined as the set of vectors that are orthogonal to all columns of $\mathbf{H}_b$:
\begin{equation}
\text{Null}(\mathbf{H}_b) = \{\mathbf{x} \in \mathbb{R}^{d} \mid \mathbf{x}^{\top}\mathbf{H}_b = \mathbf{0}\}.
\end{equation}
Constraining $\mathbf{\Delta}$ to this null space ensures that steering affects only harmful activations, while leaving benign representations invariant, thereby achieving utility preservation.

To satisfy the benign constraint in Eq.~\eqref{eq:benign-constraint}, 
we construct a null-space projection matrix $\mathbf{P}$ to restrict the transformation $\mathbf{\Delta}$ to operate only within the subspace orthogonal to benign activations. 
Following the principle in~\cite{alphaedit, alphasteer, WangL0X21}, we formulate the transformation as:
\begin{equation}\label{eq:delta}
\mathbf{\Delta} = \tilde{\mathbf{\Delta}}\mathbf{P},
\end{equation}
where $\tilde{\mathbf{\Delta}}$ is an unconstrained learnable matrix and $\mathbf{P}$ denotes the projection onto $\text{Null}(\mathbf{H}_b)$. 
This formulation naturally enforces $\mathbf{\Delta}\mathbf{H}_b = \tilde{\mathbf{\Delta}}\mathbf{P}\mathbf{H}_b = \mathbf{0}$, ensuring that benign activations remain invariant while steering acts only within non-benign directions. 

However, directly computing $\mathbf{P}$ from $\mathbf{H}_b$ can be computationally expensive since $\mathbf{H}_b \in \mathbb{R}^{d \times N_b}$ usually contains many benign samples. 
To reduce complexity, we use the fact that the null space of a matrix is identical to that of its non-central covariance matrix:
\begin{equation}\label{eq:cov_mat}
\text{Null}(\mathbf{H}_b) = \text{Null}(\mathbf{H}_b\mathbf{H}_b^{\top}).
\end{equation}
which holds because both matrices share the same left null space under rank preservation (the detailed proof is provided in Section~\ref{appx:proof} of the Supplementary
). 
This equivalence allows us to compute $\mathbf{P}$ using the much smaller matrix $\mathbf{H}_b\mathbf{H}_b^{\top} \in \mathbb{R}^{d \times d}$ when $d \ll N_b$, reducing computational cost. We then perform singular value decomposition (SVD) on $\mathbf{H}_b\mathbf{H}_b^{\top}$ to obtain:
\begin{equation}
\mathbf{H}_b\mathbf{H}_b^{\top} = \mathbf{U}\mathbf{\Lambda}\mathbf{U}^{\top},
\end{equation}
where $\mathbf{U} \in \mathbb{R}^{d \times d}$ denotes orthonormal eigenvectors, and $\mathbf{\Lambda} \in \mathbb{R}^{d \times d}$ is a diagonal matrix containing the eigenvalues. 
Since directions with zero or near-zero eigenvalues correspond to the null space of $\mathbf{H}_b$,
we select $r$ eigenvectors to form $\hat{\mathbf{U}} \in \mathbb{R}^{d \times r}$, which spans the null space of $\mathbf{H}_b$.
The projection matrix onto this space is then given by:
\begin{equation}
\mathbf{P} = \hat{\mathbf{U}}\hat{\mathbf{U}}^{\top},
\end{equation}
which satisfies $\mathbf{P}\mathbf{H}_b = \mathbf{0}$ and $\mathbf{P}^2 = \mathbf{P}$. 
This projection ensures that any transformation $\mathbf{\Delta}$ constrained as $\tilde{\mathbf{\Delta}}\mathbf{P}$ only operates within the null space of benign activations, thus preserving their representations. 
Formally, $\tilde{\mathbf{\Delta}}\mathbf{P}\mathbf{H}_b = \mathbf{0}$, as $\text{Null}(\mathbf{H}_b) = \text{Null}(\mathbf{H}_b\mathbf{H}_b^{\top})$. 
Under this null-space constraint, the steering modification vanishes for benign prompts, ensuring that the process described in Eq.~\eqref{eq:steering} leaves safe activations unchanged while selectively modulating harmful ones.

\subsection{Activation steering via null-space projection}
To guide the model away from jailbreak behaviors, we first identify latent feature directions that most contribute to harmful responses. 
Let $\mathbf{H}_{m} \in \mathbb{R}^{d \times N_m}$ denote the hidden activations collected from $N_m$ malicious prompts. 
Our goal is to learn a transformation $\mathbf{\Delta}$ that shifts these activations toward a predefined refusal direction
while ensuring that benign activations remain invariant. 
Under the null-space constraint introduced in Eq.~\eqref{eq:delta}, the transformation is expressed as $\mathbf{\Delta} = \tilde{\mathbf{\Delta}}\mathbf{P}$, 
where $\mathbf{P}$ is the projection matrix that suppresses updates within the benign subspace. The steering constraint can thus be formulated as:
\begin{equation}
\mathbf{\Delta}\mathbf{H}_{m} = \tilde{\mathbf{\Delta}}\mathbf{P}\mathbf{H}_{m} = \mathbf{R},
\end{equation}
where $\mathbf{R} \in \mathbb{R}^{d \times N_m}$ represents the target refusal activations, 
each column corresponding to the hidden state when the model generates a safe refusal for the same adversarial image under different queries. 
The basic objective aims to align steered activations with $\mathbf{R}$ while regularizing the transformation magnitude:
\begin{equation}
\tilde{\mathbf{\Delta}}^{\star} =
\arg\min_{\tilde{\mathbf{\Delta}}}
\Big(
\|\tilde{\mathbf{\Delta}}\mathbf{P}\mathbf{H}_{m} - \mathbf{R}\|_{F}^{2}
+ \alpha\|\tilde{\mathbf{\Delta}}\mathbf{P}\|_{F}^{2}
\Big),
\end{equation}
where $\alpha$ controls the smoothness of the transformation within the constrained subspace.

To further suppress residual jailbreak semantics, we extract attribution-based harmful directions $\mathbf{V}$ following the same set of malicious activations $\mathbf{H}_m$. 
For each sample $v$ in the harmful set $\mathcal{D}_m$, we mask visually salient tokens and measure the resulting activation change:
\begin{equation}\label{eq:attr_ablation}
\boldsymbol{\delta}(v) \;=\; \mathbf{h}(v) - \mathbf{h}(\mathrm{Mask}(v)) \in \mathbb{R}^{d}.
\end{equation}
Stacking these deltas across all samples yields:
\begin{equation}
\mathbf{V} \;=\; \big[\, \boldsymbol{\delta}(v_1)\; \boldsymbol{\delta}(v_2)\; \cdots\; \boldsymbol{\delta}(v_{N_m}) \,\big] \in \mathbb{R}^{d \times N_m},
\end{equation}
which captures harmful directions at the steering and serves as the suppression target in the final objective:
\begin{equation}\label{eq:final_obj}
\begin{aligned}    
\tilde{\mathbf{\Delta}}^{\star} =
\arg\min_{\tilde{\mathbf{\Delta}}}
\Big(
&\|\tilde{\mathbf{\Delta}}\mathbf{P}\mathbf{H}_{m} - \mathbf{R}\|_{F}^{2}
+ \alpha\|\tilde{\mathbf{\Delta}}\mathbf{P}\|_{F}^{2} \\
&+ \beta\|\tilde{\mathbf{\Delta}}\mathbf{P}\mathbf{H}_{m} - \mathbf{V}\|_{F}^{2}
\Big),
\end{aligned}
\end{equation}
where $\beta$ balances the suppression of harmful activation alignment.
The second term enforces stability, while the third term explicitly discourages projection onto jailbreak-related directions. The closed-form solution is derived as:
\begin{equation}\label{eq:solution}
\tilde{\mathbf{\Delta}}^{\star} 
= (\mathbf{R}+\beta\mathbf{V})\mathbf{H}_{m}^{\top}\mathbf{P}^{\top}
\big(
\mathbf{P}\mathbf{H}_{m}\mathbf{H}_{m}^{\top}\mathbf{P}^{\top}
+ (\alpha+\beta)\mathbf{P}\mathbf{P}^{\top}
\big)^{+},
\end{equation}
where $^{+}$ denotes the Moore–Penrose pseudoinverse (see the derivation in Section~\ref{appx:proof} of the Supplementary). 
At inference, each hidden activation $\mathbf{h}^{(l)}$ is updated as
\begin{equation}
\mathbf{h}^{(l)'} = 
\mathbf{h}^{(l)} + 
\lambda\,\tilde{\mathbf{\Delta}}^{\star(l)}\mathbf{P}^{(l)}\mathbf{h}^{(l)},
\end{equation}
where $\lambda$ controls the steering intensity. 
By combining null-space projection with refusal-oriented transformation, 
\texttt{NullSteer} selectively suppresses harmful activations while preserving benign behaviors.

\section{Experiments}
\label{sec:experiments}
In this section, we evaluate NullSteer on multimodal safety benchmarks, analyzing its robustness, utility preservation, and the contribution of each component through ablation and visualization.

\subsection{Setup}

\noindent{\textbf{Dataset.}}
We construct test sets for two attack scenarios, \textit{Toxicity} and \textit{Jailbreak}, using perturbation-based adversarial images. 
Following the data preparation of ASTRA~\cite{ASTRA}, we start from clean images sampled from ImageNet~\cite{imagenet} and apply the PGD attack~\cite{PGD-attack} to produce adversarial examples for both steering-vector construction and evaluation. 
Specifically, 55 benign images are sampled, from which 30 adversarial samples are used for testing, while an additional 16 adversarial images (under the same PGD settings) are generated for steering-vector construction. 
The perturbation radius is set to $\epsilon \in \{\frac{16}{255}, \frac{32}{255}, \frac{64}{255}, \text{unconstrained}\}$. 
Details of the PGD configuration are provided in Section~\ref{appx:imp_details} of the Supplementary Material.  

For textual prompts, the Toxicity test set contains 100 samples drawn from RealToxicityPrompt~\cite{RealToxicityPrompts}, while the Jailbreak test set includes 110 prompts sampled from Advbench~\cite{Advbench} and Anthropic-HHH~\cite{Anthropic-HHH}. 
All textual prompts are disjoint from those used in steering-vector construction. 
We evaluate model utility in benign settings using two widely adopted multimodal reasoning benchmarks, MM-Vet~\cite{MM-Vet} and MM-Bench~\cite{MMBench}, which cover diverse vision-language understanding and instruction-following tasks. 
To additionally examine over-refusal behaviors, we include safe instructions from XSTest~\cite{XSTest}, allowing analysis of whether safety mechanisms affect normal task responses. 

\noindent{\textbf{Models \& Baselines.}} 
We conduct experiments on three representative open-source VLMs, including Qwen2-VL-7B~\cite{Qwen-VL}, MiniGPT-4-13B~\cite{MiniGPT-4}, and LLaVA-v1.5-13B~\cite{llava-v1.5}. 
The steering layer $l$ is set to 20 for 13B models and 14 for 7B models. 
We follow the experimental settings of previous work~\cite{ASTRA} for attribution ablation in Eq.~\eqref{eq:attr_ablation}. 
For comparison, we include three VLM defense baselines and three steering-based approaches. 
Among VLM defenses, Self-Reminder~\cite{self-reminders} introduces safety control via system prompts, 
JailGuard~\cite{JailbreakBench} detects unsafe content by perturbing input images and comparing response divergence, 
and ECSO~\cite{ECSO} converts unsafe visual inputs into text to activate intrinsic safety mechanisms of the underlying LLM. 
For steering-based defenses, Wang et al.~\cite{ASTRA} perform adaptive activation steering by deriving transferable vectors from ablated adversarial tokens, and Rimsky et al.~\cite{Rimsky} with Ball et al.~\cite{Ball} construct textual steering vectors encoding refusal semantics and jailbreak templates.

\subsection{Main results}

\begin{table*}[t!]
\tabstyle{2pt}
\centering
\caption{
Defense comparisons across different VLMs. 
Lower values ($\downarrow$) indicate stronger robustness. 
Steering vectors for each $\epsilon$ are derived from adversarial examples generated under the same perturbation level. The best results are highlighted in \textbf{bold}.
}
\label{tab:combined_defense_rotated}
\begin{tabular}{cc|cccc|cccc}
\toprule
 \multicolumn{2}{c}{\multirow{3}{*}{Method}} 
 & \multicolumn{4}{|c|}{Toxicity (Perturbation-based) -- Toxicity Score (\%) $~\downarrow$}
 & \multicolumn{4}{c}{Jailbreak (Perturbation-based) -- ASR (\%) $~\downarrow$} \\
\cmidrule(lr){3-10}
 & & $\epsilon={16}/{255}$ & $\epsilon={32}/{255}$ & $\epsilon={64}/{255}$ & unconstrained 
   & $\epsilon={16}/{255}$ & $\epsilon={32}/{255}$ & $\epsilon={64}/{255}$ & unconstrained \\
\midrule
\multirow{9}{*}{\rotatebox{90}{MiniGPT-4}} 
 & \multicolumn{1}{|c|}{Benign image} & 30.65 & 30.65 & 30.65 & 30.65 & 24.55 & 24.55 & 24.55 & 24.55 \\
& \multicolumn{1}{|c|}{w/o defense} & 39.73 & 48.52 & 54.70 & 52.12 & 44.55 & 47.27 & 49.09 & 53.64 \\
& \multicolumn{1}{|c|}{Self-reminder~\cite{self-reminders}} & 38.97 & 48.71 & 45.15 & 50.12 & 35.45 & 36.36 & 41.82 & 43.64 \\
& \multicolumn{1}{|c|}{JailGuard~\cite{JailGuard}} & 16.51 & 18.93 & 20.93 & 21.23 & 30.00 & 32.73 & 27.27 & 28.18 \\
& \multicolumn{1}{|c|}{ECSO~\cite{ECSO}} & 34.59 & 32.42 & 38.54 & 42.86 & 40.91 & 42.73 & 29.09 & 37.27 \\
& \multicolumn{1}{|c|}{Refusal Pairs~\cite{Refusal_Pairs}}& 25.76 & 30.28 & 31.99 & 35.71 & 20.00 & 22.73 & 17.27 & 16.36 \\
& \multicolumn{1}{|c|}{Jailbreak Templates~\cite{Jailbreak_Templates}} & 19.73 & 25.03 & 30.10 & 22.78 & 33.64 & 38.15 & 38.18 & 42.73 \\
& \multicolumn{1}{|c|}{ASTRA~\cite{ASTRA}} & {11.30} & {8.84} & {4.51} & {4.48} & {9.09} & {13.18} & {15.46} & {9.09} \\
& \multicolumn{1}{|c|}{\texttt{NullSteer}} & \textbf{9.52} & \textbf{6.45} & \textbf{3.26} & \textbf{2.89} & \textbf{7.62} & \textbf{10.72} & \textbf{13.58} & \textbf{7.32} \\
\midrule
\multirow{9}{*}{\rotatebox{90}{Qwen2-VL}} 
& \multicolumn{1}{|c|}{Benign image} & 38.52 & 38.52 & 38.52 & 38.52 & 0.00 & 0.00 & 0.00 & 0.00 \\
& \multicolumn{1}{|c|}{w/o defense} & 50.50 & 51.62 & 55.59 & 53.43 & 67.27 & 70.46 & 71.82 & 76.36 \\
& \multicolumn{1}{|c|}{Self-reminder~\cite{self-reminders}} & 30.47 & 27.53 & 32.84 & 29.09 & 50.00 & 47.27 & 40.00 & 58.18 \\
& \multicolumn{1}{|c|}{JailGuard~\cite{JailGuard}} & 29.37 & 24.68 & 28.74 & 27.76 & 19.09 & 20.00 & 21.82 & {15.45} \\
& \multicolumn{1}{|c|}{ECSO~\cite{ECSO}} & 50.09 & 50.68 & 56.08 & 51.57 & 30.00 & 27.27 & 31.82 & 32.73 \\
& \multicolumn{1}{|c|}{Refusal Pairs~\cite{Refusal_Pairs}} & 46.14 & 46.83 & 46.83 & 40.53 & 29.09 & 31.82 & 21.82 & 52.73 \\
& \multicolumn{1}{|c|}{Jailbreak Templates~\cite{Jailbreak_Templates}} & 66.74 & 63.35 & 67.15 & 68.29 & 68.18 & 68.18 & 65.45 & 74.55 \\
& \multicolumn{1}{|c|}{ASTRA~\cite{ASTRA}} & 15.52 & 5.45 & 2.39 & 0.07 & 6.06 & 5.00 & 18.18 & 15.45 \\
& \multicolumn{1}{|c|}{\texttt{NullSteer}} & \textbf{13.36} & \textbf{3.51} & \textbf{1.98} & \textbf{0.05} & \textbf{4.21} & \textbf{4.55} & \textbf{15.82} & \textbf{13.25} \\
\midrule
\multirow{9}{*}{\rotatebox{90}{LLaVA-v1.5}} 
& \multicolumn{1}{|c|}{Benign image} & 75.00 & 75.00 & 75.00 & 75.00 & 13.64 & 13.64 & 13.64 & 13.64 \\
& \multicolumn{1}{|c|}{w/o defense} & 83.70 & 84.40 & 85.54 & 85.44 & 51.82 & 56.36 & 55.45 & 53.64 \\
& \multicolumn{1}{|c|}{Self-reminder~\cite{self-reminders}} & 83.92 & 83.97 & 84.19 & 80.93 & 28.18 & 30.00 & 22.73 & 22.73 \\
& \multicolumn{1}{|c|}{JailGuard~\cite{Jailbreak_Templates}} & 77.60 & 77.77 & 75.76 & 73.76 & 23.64 & 21.82 & 30.00 & 17.27 \\
& \multicolumn{1}{|c|}{ECSO~\cite{ECSO}} & 73.77 & 73.14 & 71.32 & 66.81 & 24.55 & 21.82 & 14.55 & 20.00 \\
& \multicolumn{1}{|c|}{Refusal Pairs~\cite{Refusal_Pairs}} & 66.72 & 66.82 & 60.36 & 62.46 & 23.64 & 25.45 & 20.00 & 19.09 \\
& \multicolumn{1}{|c|}{Jailbreak Templates~\cite{Jailbreak_Templates}} & 52.61 & 50.21 & 55.48 & 54.90 & 23.64 & 17.27 & 20.00 & 29.09 \\
& \multicolumn{1}{|c|}{ASTRA~\cite{ASTRA}} & {36.02} & {34.76} & {43.13} & {25.10} & {4.55} & {10.91} & {13.64} & {12.43} \\
& \multicolumn{1}{|c|}{\texttt{NullSteer}} & \textbf{32.55} & \textbf{31.82} & \textbf{39.32} & \textbf{22.35} & \textbf{3.18} & \textbf{8.75} & \textbf{11.48} & \textbf{10.25} \\
\bottomrule
\end{tabular}
\end{table*}

\noindent{\textbf{Defense performance.}} 
Table~\ref{tab:combined_defense_rotated} reports the defense results under perturbation-based attacks on MiniGPT-4, Qwen2-VL, and LLaVA-v1.5. 
Across all settings, \textit{NullSteer} consistently attains the lowest Toxicity Score and ASR, indicating stable robustness and generalization across architectures. 
Under the strongest unconstrained attack on MiniGPT-4, \texttt{NullSteer} records a Toxicity Score of 2.89\% and an ASR of 7.32\%, compared with 42.86\% and 37.27\% from ECSO and 4.48\% and 9.09\% from ASTRA. 
Similar trends are observed on Qwen2-VL and LLaVA-v1.5. 
For Qwen2-VL under $\epsilon{=}32/255$, \texttt{NullSteer} attains a Toxicity of 3.51\% and an ASR of 4.55\%, while ASTRA~\cite{ASTRA} reports 5.45\% and 5.00\%, respectively. 
For LLaVA-v1.5 under $\epsilon{=}32/255$, \texttt{NullSteer} achieves 31.82\% Toxicity and 8.75\% ASR, compared with 34.76\% and 10.91\% from ASTRA. 
Traditional VLM defenses such as Self-Reminder~\cite{self-reminders}, JailGuard~\cite{JailbreakBench}, and ECSO~\cite{ECSO} remain highly vulnerable, typically exhibiting Toxicity above 40\% and ASR exceeding 50\%, highlighting the limitations of input- or output-level adjustments. 
In contrast, steering-based defenses provide stronger alignment control, yet Refusal Pairs~\cite{Refusal_Pairs} and Jailbreak Templates~\cite{Jailbreak_Templates} show considerably higher Toxicity and ASR due to the lack of multimodal coupling. 
Compared with ASTRA, \texttt{NullSteer} consistently achieves lower scores across all benchmarks, demonstrating the advantage of null-space–constrained activation steering in enhancing robustness and safety simultaneously. 

\begin{figure*}[!t]
\centering
\begin{minipage}{0.49\textwidth}
    \centering
    \begin{subfigure}[b]{0.49\textwidth}
        \centering
        \includegraphics[width=\textwidth]{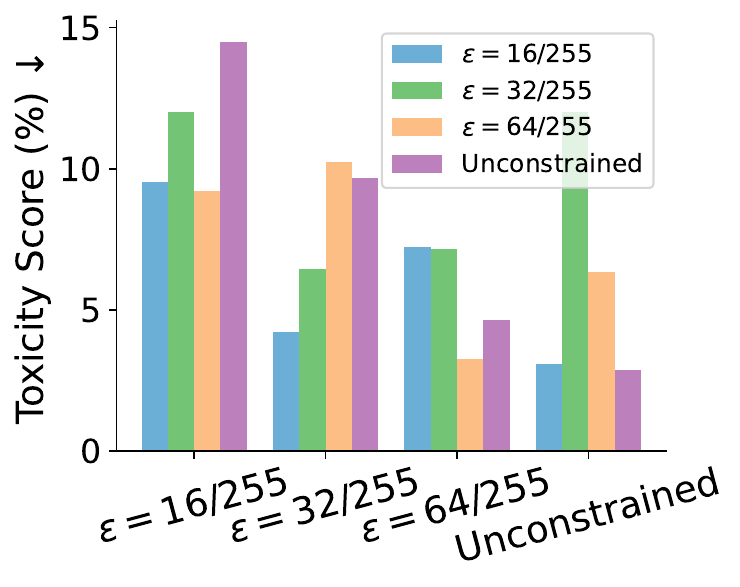}
        \caption{Toxicity on MiniGPT-4}
        \label{fig:id_tox_minigpt4}
    \end{subfigure}
    \hfill
    \begin{subfigure}[b]{0.49\textwidth}
        \centering
        \includegraphics[width=\textwidth]{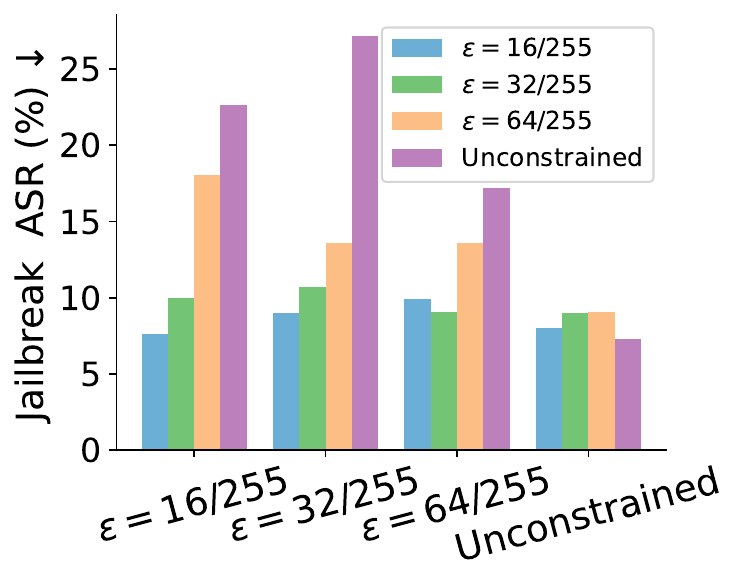}
        \caption{Jailbreak on MiniGPT-4}
        \label{fig:id_jai_minigpt4}
    \end{subfigure}
    \caption{Transferability performance under ID conditions.}
    \label{fig:id_minigpt4}
\end{minipage}
\hfill
\begin{minipage}{0.49\textwidth}
    \centering
    \begin{subfigure}[b]{0.49\textwidth}
        \centering
        \includegraphics[width=\textwidth]{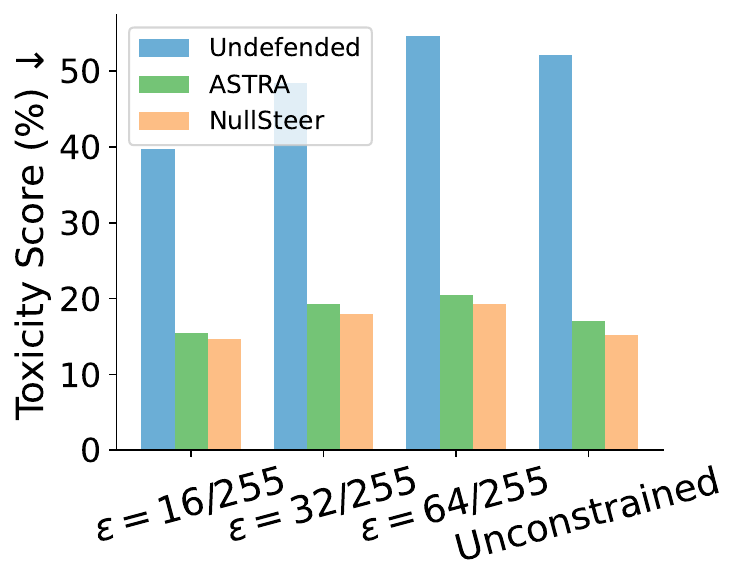}
        \caption{Adaptive Toxicity on MiniGPT-4}
        \label{fig:ada_tox_minigpt4}
    \end{subfigure}
    \hfill
    \begin{subfigure}[b]{0.49\textwidth}
        \centering
        \includegraphics[width=\textwidth]{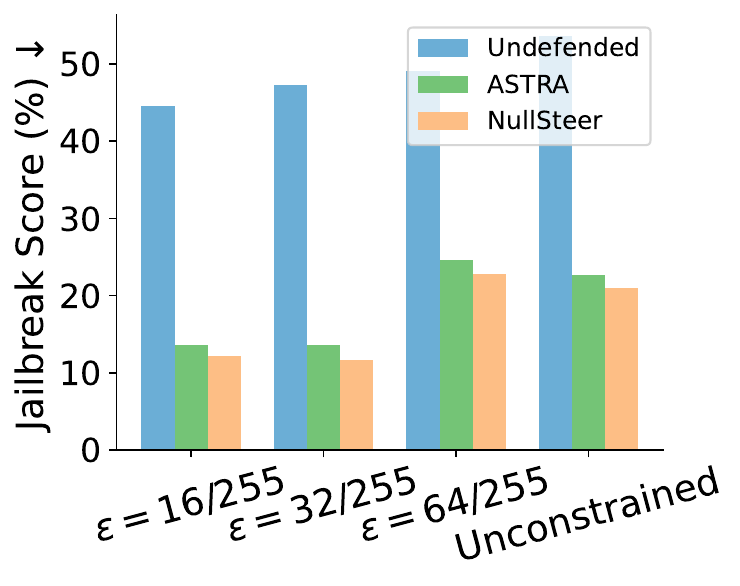}
        \caption{Adaptive Jailbreak on MiniGPT-4}
        \label{fig:ada_jai_minigpt4}
    \end{subfigure}
    \caption{Adaptive attack performance on MiniGPT-4.}
    \label{fig:ada_minigpt4}
\end{minipage}
\end{figure*}

In the in-distribution (ID) setting, steering vectors are constructed and evaluated on adversarial samples from the same ImageNet classes to assess robustness across perturbation intensities. As shown in Figure~\ref{fig:id_minigpt4}, \texttt{NullSteer} maintains low Toxicity and Jailbreak scores across varying $\epsilon$ levels. The defense derived from $\epsilon=\tfrac{16}{255}$ remains effective against $\epsilon=\tfrac{64}{255}$ attacks, with only 7.3\% Toxicity and 9.0\% Jailbreak. On average, \texttt{NullSteer} achieves 5.78\% Toxicity and 3.00\% Jailbreak, improving over ASTRA by 4–5\%. Overall, these results confirm that \texttt{NullSteer} encodes robust, transferable steering directions that generalize both across and within perturbation distributions.
To further validate the generalization of \texttt{NullSteer} beyond in-distribution settings, we additionally perform out-of-distribution (OOD) experiments, with results presented in Section~\ref{appx:more_results} of the Supplementary Material.

\noindent{\textbf{Adaptive attack performance.}} Adaptive attack assumes the adversary has full knowledge of the defense, including the steering vectors and coefficients. Following this setting, we perform PGD-based adaptive attacks under different perturbation magnitudes. As shown in Figure~\ref{fig:ada_minigpt4}, both ASTRA and our \texttt{NullSteer} reduce Toxicity and Jailbreak scores compared to the undefended baseline. Notably,  our \texttt{NullSteer} achieves the lowest Jailbreak rate across all $\epsilon$ values, e.g., reducing it from 49.1\% to 19.3\% at $\epsilon=\tfrac{64}{255}$. These results demonstrate that \texttt{NullSteer} maintains strong robustness even when the attacker optimizes against the defense, validating its stability under the most challenging white-box setting.

\begin{table*}[!ht]
\tabstyle{2pt}
\centering
\caption{Utility performance in benign and adversarial scenarios.}
\begin{tabular}{ll|ccc|ccc}
\toprule
\multicolumn{2}{c|}{\multirow{3.5}{*}{Methods}} 
& \multicolumn{3}{c|}{Benign Scenarios -- Utility Score $~\uparrow$} 
& \multicolumn{3}{c}{Adversarial Scenarios -- Perplexity $~\downarrow$} \\ 
\cmidrule(lr){3-8}
& 
& MM-Vet~\cite{MM-Vet} 
& MMBench~\cite{MMBench} 
& XSTest~\cite{XSTest}
& \makecell[c]{Toxicity \\ (Perturbation-based)} 
& \makecell[c]{Jailbreak \\ (Perturbation-based)} 
& \makecell[c]{Jailbreak \\ (Structured-based)} \\
\midrule

\multirow{3}{*}{MiniGPT-4} 
& \multicolumn{1}{|c|}{Vanilla} & 19.40 & 35.90 & 87.60 & 51.42 & 3.95 & 2.62 \\
& \multicolumn{1}{|c|}{ASTRA~\cite{ASTRA}}   & 20.62 & 35.82 & 87.60 & 10.14 & 5.82 & 4.29 \\
& \multicolumn{1}{|c|}{\texttt{NullSteer}}  & \textbf{21.05} & \textbf{36.25} & \textbf{87.80} & \textbf{9.52} & \textbf{2.58} & \textbf{3.65} \\
\midrule

\multirow{3}{*}{LLaVA-v1.5} 
& \multicolumn{1}{|c|}{Vanilla} & \textbf{32.62} & 72.94 & 98.00 & 63.68 & 3.68 & 3.82 \\
& \multicolumn{1}{|c|}{ASTRA~\cite{ASTRA}}   & 30.55 & 73.23 & \textbf{98.80} & 59.28 & 8.59 & 4.61 \\
& \multicolumn{1}{|c|}{\texttt{NullSteer}}  & 31.20 & \textbf{73.56} & \textbf{98.80} & \textbf{57.96} & \textbf{7.23} & \textbf{4.26} \\
\midrule

\multirow{3}{*}{Qwen2-VL} 
& \multicolumn{1}{|c|}{Vanilla} & 49.13 & 78.00 & 73.60 & 140.44 & \textbf{6.80} & \textbf{30.00} \\
& \multicolumn{1}{|c|}{ASTRA~\cite{ASTRA}}  & 48.66 & 78.69 & 74.00 & 40.14 & 8.86 & 30.92 \\
& \multicolumn{1}{|c|}{\texttt{NullSteer}}   & \textbf{49.02} & \textbf{78.82} & \textbf{74.50} & \textbf{38.45} & 6.95 & 30.48 \\
\bottomrule
\end{tabular}%
\label{tab:utility}
\end{table*}

\noindent{\textbf{Utility performance.}} Table~\ref{tab:utility} compares the utility of defended models under benign and adversarial conditions. 
Across MM-Vet, MMBench, and XSTest, \texttt{NullSteer} maintains comparable or higher utility scores than ASTRA and Vanilla models, indicating that our defense does not compromise benign performance. 
For instance, on MiniGPT-4, the MM-Vet score slightly improves from 19.40 to 21.05, and similar trends are observed on Qwen2-VL and LLaVA-v1.5. 
Under adversarial scenarios, \texttt{NullSteer} consistently achieves the lowest perplexity, confirming its ability to produce coherent and safe outputs when facing harmful prompts. 
On MiniGPT-4, the toxicity perplexity drops from 51.42 to 9.52, and the jailbreak perplexity from 3.95 to 2.58, demonstrating strong robustness without degrading response quality. 
This balance between safety and utility stems from the adaptive steering mechanism, which modulates activation shifts in proportion to semantic deviation, preventing over-suppression of normal responses while effectively reducing harmful generation.

\begin{table*}[t]
\tabstyle{5pt}
\centering
\caption{
Ablation study of different objective components on Minigpt-4.
}
\begin{tabular}{ccc|ccc}
\toprule
$\|\tilde{\mathbf{\Delta}}\mathbf{P}\|_{F}^{2}$ 
& $\|\tilde{\mathbf{\Delta}}\mathbf{P}\mathbf{H}_{m} - \mathbf{R}\|_{F}^{2}$ 
& $\|\tilde{\mathbf{\Delta}}\mathbf{P}\mathbf{H}_{m} - \mathbf{V}\|_{F}^{2}$ 
& Toxicity Score $\downarrow$ 
& Jailbreak (ASR) $\downarrow$ 
& Utility Score $\uparrow$ \\ 
\midrule

\XSolidBrush & \XSolidBrush & \XSolidBrush & 30.65 & 34.55 & 35.90 \\
\Checkmark   & \Checkmark   & \XSolidBrush & 3.58  & 8.36  & 36.00 \\
\Checkmark   & \XSolidBrush & \Checkmark   & 4.02  & 8.57  & 36.00 \\
\Checkmark   & \Checkmark   & \Checkmark   & \textbf{2.89} & \textbf{7.32} & \textbf{36.25} \\
\bottomrule
\end{tabular}
\label{tab:ablation_module}
\end{table*}

\subsection{Ablation study}
\noindent{\textbf{Objective function analysis.}}  
We analyze the contribution of the three loss components in Eq.~\eqref{eq:final_obj}: the smoothness regularization 
$\|\tilde{\Delta}\mathbf{P}\|_F^2$, the refusal alignment 
$\|\tilde{\Delta}\mathbf{P}\mathbf{H}_m - \mathbf{R}\|_F^2$, and the harmful suppression 
$\|\tilde{\Delta}\mathbf{P}\mathbf{H}_m - \mathbf{V}\|_F^2$. 
As shown in Table~\ref{tab:ablation_module}, under the unconstrained perturbation setting on MiniGPT-4, removing any term notably weakens the defense, confirming their complementary effects. 
The refusal-alignment term provides the dominant safety improvement by steering harmful activations toward refusal semantics, 
while the harmful-suppression term further reduces residual correlations with jailbreak features, enhancing robustness under strong perturbations. 
The smoothness regularization constrains transformation magnitude, ensuring stability in benign subspaces and preserving high utility. 
When all three objectives are combined, \texttt{NullSteer} attains the lowest Toxicity (2.89\%) and ASR (7.32\%) with the highest MMBench score (36.25), 
achieving a balanced trade-off between safety and utility.

\begin{figure}[t]
  \centering
 \begin{subfigure}[b]{0.49\linewidth}
        \centering
        \includegraphics[width=\linewidth]{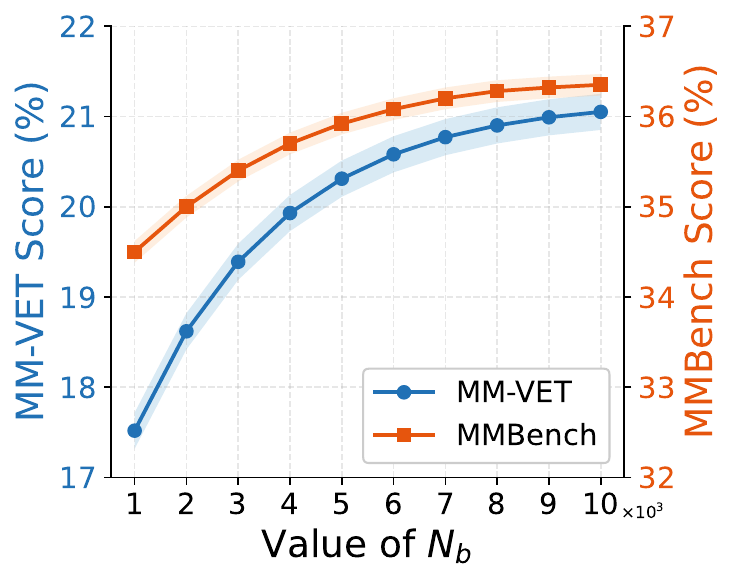}
        \caption{MiniGPT-4}
        \label{fig:ig:utility_para_minigpt4}
    \end{subfigure}
    \hfill
    \begin{subfigure}[b]{0.49\linewidth}
        \centering
        \includegraphics[width=\linewidth]{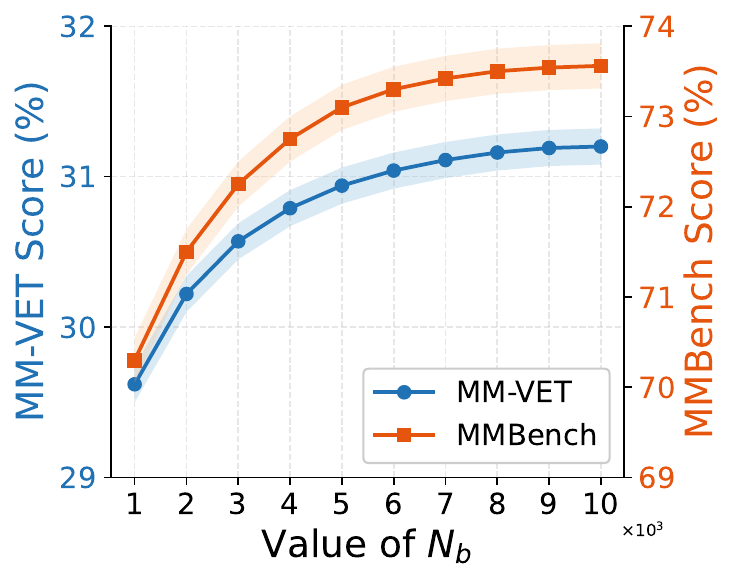}
        \caption{LLaVA-V1.5}
        \label{fig:utility_para_llava}
    \end{subfigure}
\caption{{Utility with varying numbers of benign activations $N_b$.}
Evaluation is conducted on MM-VET and MMBench with MiniGPT-4 and LLaVA-v1.5 
under the{unconstrained}.}
\label{fig:utility_para_num}
\end{figure}
\noindent{\textbf{Analysis of the number of benign activations.}}
As shown in Figure~\ref{fig:utility_para_num}, increasing the number of benign activations $N_b$ used to estimate the null space consistently improves model utility on both MM-VET and MMBench under the unconstrained perturbation setting.
When $N_b$ is small, the estimated null space is under-constrained, leading to minor degradation in benign performance.
As $N_b$ increases, the projection matrix becomes more stable and accurately captures the invariant subspace of benign activations, resulting in steady utility gains for both MiniGPT-4 and LLaVA-v1.5.
The improvement gradually saturates around $N_b\!\approx\!8$, indicating that only a moderate number of benign samples is sufficient to construct a well-conditioned projection without additional computation overhead.

\begin{figure}
    \label{fig:placeholder}
    \begin{subfigure}[b]{0.49\linewidth}
        \centering
        \includegraphics[width=\linewidth]{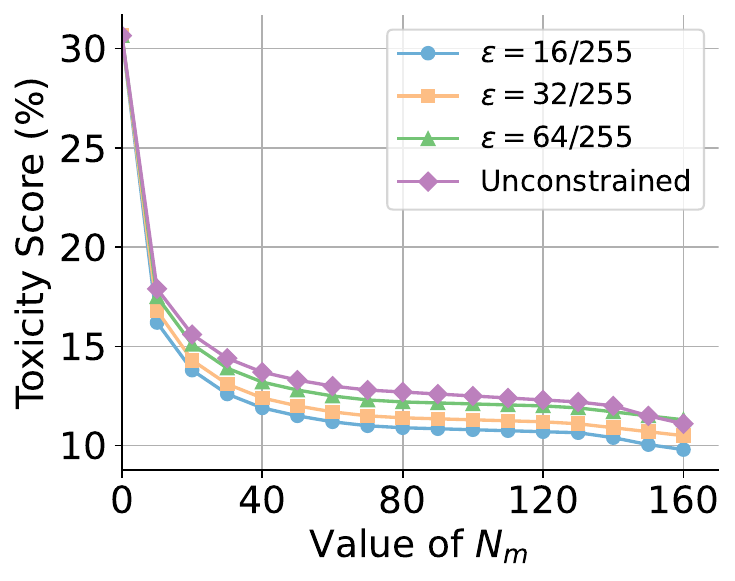}
        \caption{Toxicity score on MiniGPT-4}
        \label{fig:tox_num_minigpt4}
    \end{subfigure}
    \hfill
    \begin{subfigure}[b]{0.49\linewidth}
        \centering
        \includegraphics[width=\linewidth]{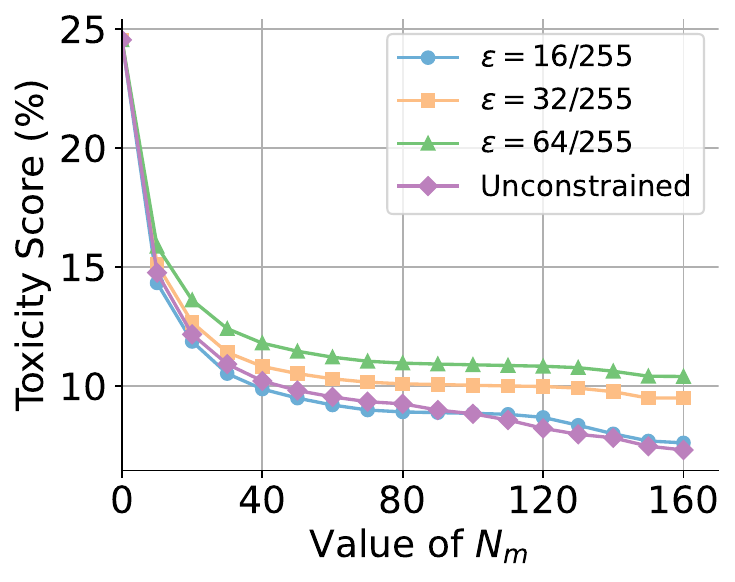}
        \caption{Jailbreak ASR on MiniGPT-4.}
        \label{fig:jai_num_minigpt4}
    \end{subfigure}
    \caption{{Safety with varying numbers of malicious activations $N_m$.}
    Evaluation is conducted on MiniGPT-4 under multiple perturbation levels.}
    \label{fig:safety_num_minigpt4}
\end{figure}

\noindent{\textbf{Analysis of the number of malicious activations.}} 
As shown in Figure~\ref{fig:safety_num_minigpt4}, increasing the number of malicious activations $N_m$ used to estimate harmful directions improves the defense effectiveness across all perturbation levels.
When $N_m$ is small, the estimated harmful subspace fails to capture consistent jailbreak semantics, leading to higher Toxicity and ASR.
As $N_m$ increases, both metrics rapidly decline and stabilize, suggesting that richer harmful evidence enables more reliable direction estimation.
Beyond $N_m\!\approx\!100$, the performance gain becomes marginal, indicating that a moderate number of malicious samples is sufficient for  efficient steering vector construction.

\begin{figure}[t]
  \centering
  \begin{subfigure}{0.49\linewidth}
  \centering
    \includegraphics[width=1\linewidth]{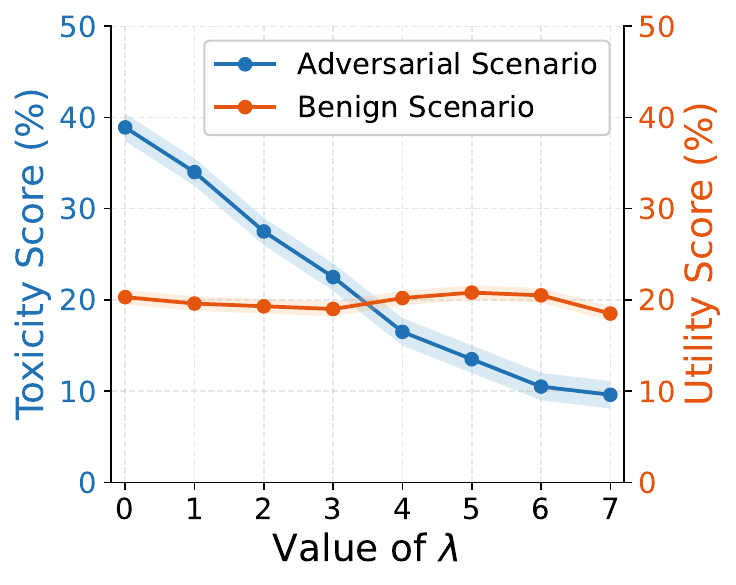}
    \caption{Toxicity Attack}
    \label{fig:steer_strenght_tox}
  \end{subfigure}
 \begin{subfigure}{0.49\linewidth}
    \centering
    \includegraphics[width=1\linewidth]{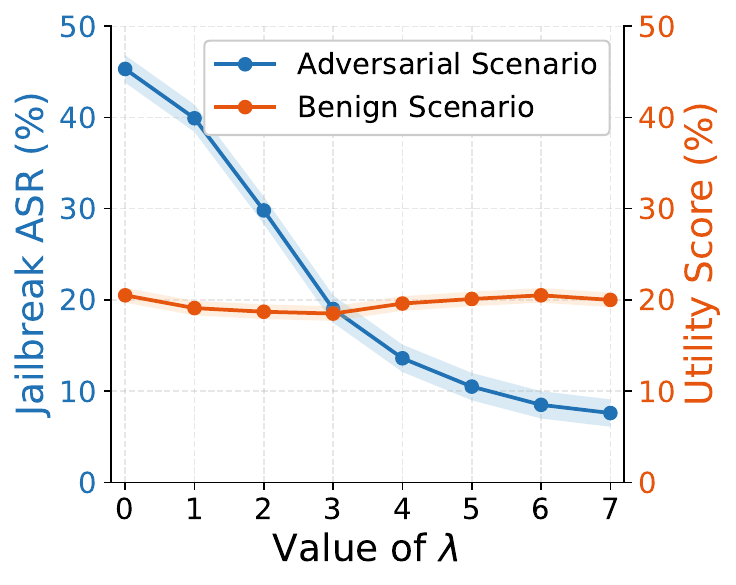}
    \caption{Jailbreak Attack}
    \label{fig:steer_strength_jai}
  \end{subfigure}
  \caption{Effect of steering strength on MiniGPT-4. We vary $\lambda$ when applying activation steering under the $\epsilon\!=\!16/255$ perturbation setting and report Toxicity Score on the Toxicity attack, Jailbreak ASR on the Jailbreak attack, and utility on MM-Vet.}
  \label{fig:steer_strength}
\end{figure}
\noindent{\textbf{Analysis of steering strength.}} Figure~\ref{fig:steer_strength} illustrates that increasing the steering strength $\lambda$ progressively suppresses unsafe behaviors on MiniGPT-4.
Both Toxicity and Jailbreak ASR drop markedly as $\lambda$ grows, confirming that stronger activation modulation enhances refusal robustness under adversarial perturbations.
The MM-Vet utility remains stable across all settings, demonstrating that steering primarily affects adversarial directions while preserving benign representations.
An intermediate strength ($\lambda\!\approx\!5$) yields the best safety–utility balance.

\begin{table}[!t]
\tabstyle{2pt}
\centering   
\caption{{Inference time per token (ms).}
Average decoding latency is computed and normalized by the number of generated tokens. 
Reporting time per token enables a fair comparison of inference efficiency across models with varying output lengths.
Lower values indicate faster decoding.}
\begin{tabular}{c | c c c | c }
\toprule
 & MiniGPT-4 & LLaVA-v1.5 & Qwen2-VL & Avg. \\ 
\midrule
{w/o defense} & 173.19 & 40.68 & 27.43 & 63.67 \\
{Self-reminder~\cite{self-reminders}} & 173.36 & 41.09 & 27.94 & 53.38 \\
{JailGuard~\cite{Jailbreak_Templates}} & 1557.98 & 366.02 & 245.42 & 40.92 \\
{ECSO~\cite{ECSO}} & 457.55 & 116.44 & 70.22 & 53.75 \\
{ASTRA}~\cite{ASTRA} & 173.77 & 40.69 & 27.98 & 9.88 \\
\midrule
{\texttt{NullSteer}} & 173.57 & 40.68 & 27.66 & 8.43 \\ 
\bottomrule
\end{tabular}  
\label{tab:inference_time}
\end{table}
\noindent{\textbf{Inference time.}}
Table~\ref{tab:inference_time} compares the average decoding latency of different defenses.
\texttt{NullSteer} maintains comparable inference speed to the original models, requiring 173.57 ms/token on MiniGPT-4, 40.68 ms/token on LLaVA-v1.5, and 27.66 ms/token on Qwen2-VL. Its average latency of 8.43 shows negligible overhead compared with the undefended models. In contrast, multi-round or auxiliary-based defenses such as JailGuard and ECSO introduce substantial delays, exceeding 1.5 s/token and 450 ms/token, respectively. Compared with the ASTRA, \texttt{NullSteer} achieves slightly lower latency while maintaining robustness, showing its efficiency and practicality for real-time deployment.

\begin{figure}
  \centering
    \includegraphics[width=0.6\linewidth]{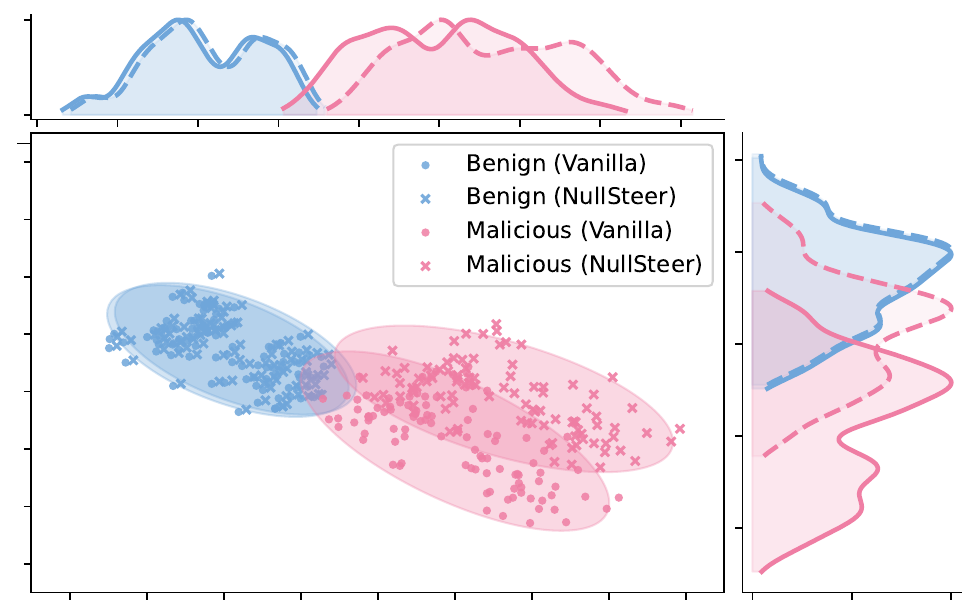}
    \caption{
    {Visualization of activation distributions.} 
    }
    \label{fig:act_distribution}
\end{figure}

\noindent{\textbf{Activation distribution analysis.}}  
Figure~\ref{fig:act_distribution} visualizes the activations distributions of benign and malicious inputs before and after applying \texttt{NullSteer} on MiniGPT-4. 
The benign clusters remain well aligned with the original representations, indicating that our null-space constraint effectively preserves utility by preventing changes in safe directions. 
In contrast, the malicious activations exhibit a clear displacement toward the refusal region, confirming that \texttt{NullSteer} adaptively redirects unsafe semantics. More visualization results are provided in Section~\ref{appx:more_results} of the Supplementary Material.

\section{Conclusion}
In this work, 
We presented \texttt{NullSteer}, a principled activation-steering framework that enhances the safety of vision–language models through null-space–constrained representation modulation. 
By restricting steering to directions orthogonal to benign activations, \texttt{NullSteer} preserves normal behavior while effectively suppressing unsafe semantics. 
Extensive evaluations demonstrate its superior robustness under both perturbation-based and adaptive attacks without compromising utility. 
Our study highlights the potential of representation-level alignment as a general and efficient paradigm for building safe and reliable multimodal large language models. 
Future work will explore extending null-space steering to larger-scale alignment tasks and integrating it with adaptive safety tuning during fine-tuning or instruction following.

\section*{Acknowledge}
This research is supported by the National Natural Science Foundation of China (No.U24B20180, No. 62576330) and National Natural Science Foundation of Anhui (No.2508085MF143).

{
    \small
    \bibliographystyle{ieeenat_fullname}
    \bibliography{main}
}

\renewcommand{\thetable}{\Alph{table}}
\renewcommand{\thefigure}{\Alph{figure}}
\clearpage
\setcounter{page}{1}
\newpage
\appendix

\section{Derivation of Eq.~\eqref{eq:cov_mat} and Eq.~\eqref{eq:solution}}\label{appx:proof}
\subsection{Eq.~\eqref{eq:cov_mat}: Null-space equivalence}
Let $\mathbf{H}_b \in \mathbb{R}^{d \times N_b}$ be the benign activation matrix.  
Eq.~\eqref{eq:cov_mat} claims that the left null space of $\mathbf{H}_b$ coincides with that of its Gram matrix:
\begin{equation}
\text{Null}(\mathbf{H}_b)=\text{Null}(\mathbf{H}_b\mathbf{H}_b^\top).
\label{eq:app_eq7_newgoal}
\end{equation}

The null spaces are defined by:
\begin{align}
\text{Null}(\mathbf{H}_b)
&=\{\mathbf{x}\in\mathbb{R}^d:\mathbf{x}^\top\mathbf{H}_b=\mathbf{0}^\top\}, \\
\text{Null}(\mathbf{H}_b\mathbf{H}_b^\top)
&=\{\mathbf{x}\in\mathbb{R}^d:\mathbf{x}^\top\mathbf{H}_b\mathbf{H}_b^\top=\mathbf{0}^\top\}.
\end{align}

To show the inclusion $\text{Null}(\mathbf{H}_b)\subseteq \text{Null}(\mathbf{H}_b\mathbf{H}_b^\top)$,  
take any $\mathbf{x}$ such that $\mathbf{x}^\top\mathbf{H}_b=\mathbf{0}^\top$.  
Using this condition:
\begin{equation}
\mathbf{x}^\top\mathbf{H}_b\mathbf{H}_b^\top
=(\mathbf{x}^\top\mathbf{H}_b)\mathbf{H}_b^\top
=\mathbf{0}^\top,
\end{equation}
which confirms $\mathbf{x}\in\text{Null}(\mathbf{H}_b\mathbf{H}_b^\top)$.
Conversely, assume $\mathbf{x}^\top\mathbf{H}_b\mathbf{H}_b^\top=\mathbf{0}^\top$.  
Since $\mathbf{H}_b\mathbf{H}_b^\top$ is symmetric and positive semidefinite, its quadratic form satisfies:
\begin{equation}
\mathbf{x}^\top\mathbf{H}_b\mathbf{H}_b^\top\mathbf{x}
=\|\mathbf{H}_b^\top\mathbf{x}\|_2^2 \ge 0.
\end{equation}
Multiplying the nullity condition by $\mathbf{x}$ yields:
\begin{equation}
0=\mathbf{x}^\top\mathbf{H}_b\mathbf{H}_b^\top\mathbf{x}
=\|\mathbf{H}_b^\top\mathbf{x}\|_2^2,
\end{equation}
which forces $\mathbf{H}_b^\top \mathbf{x}=\mathbf{0}$ and thus $\mathbf{x}^\top\mathbf{H}_b=\mathbf{0}^\top$.  
Hence $\mathbf{x}\in\text{Null}(\mathbf{H}_b)$.
Since each null space is contained in the other, the equivalence in Eq.~\eqref{eq:app_eq7_newgoal} follows:
\begin{equation}
\text{Null}(\mathbf{H}_b)=\text{Null}(\mathbf{H}_b\mathbf{H}_b^\top).
\end{equation}

\subsection{Closed-form Solution of Eq.~\eqref{eq:solution}}

We derive the closed-form solution of the regularized least-squares problem in Eq.~\eqref{eq:final_obj}.  
Recall that the objective is:
\begin{equation*}
\begin{aligned}
\tilde{\boldsymbol{\Delta}}^{\star}
=
\arg\min_{\tilde{\boldsymbol{\Delta}}}
\Big(
&\|\tilde{\boldsymbol{\Delta}}\mathbf{P}\mathbf{H}_m - \mathbf{R}\|_F^2 \\
&+ \alpha \|\tilde{\boldsymbol{\Delta}}\mathbf{P}\|_F^2
+ \beta \|\tilde{\boldsymbol{\Delta}}\mathbf{P}\mathbf{H}_m - \mathbf{V}\|_F^2
\Big),
\quad \\
&\alpha>0,\ \beta\ge 0.
\label{eq:app_obj_original}
\end{aligned}
\end{equation*}

For notational simplicity, we introduce:
\begin{equation}
\begin{aligned}
\mathbf{X} &:= \mathbf{P}\mathbf{H}_m \in \mathbb{R}^{d\times N_m},\\
\mathbf{Z} &:= \mathbf{P} \in \mathbb{R}^{d\times d}, \\
\mathbf{Y} &:= \mathbf{R} + \beta \mathbf{V} \in \mathbb{R}^{d\times N_m},\\
\mathbf{W} &:= \tilde{\boldsymbol{\Delta}} \in \mathbb{R}^{d\times d}.
\end{aligned}
\end{equation}
Grouping the two alignment terms in \eqref{eq:app_obj_original}, the objective can be written in the compact form:
\begin{equation}
J(\mathbf{W})
=
\|\mathbf{W}\mathbf{X} - \mathbf{Y}\|_F^2
+ (\alpha+\beta)\,\|\mathbf{W}\mathbf{Z}\|_F^2,
\label{eq:app_obj_compact}
\end{equation}
which corresponds to a matrix-valued ridge regression with effective target $\mathbf{Y}$ and regularization coefficient $\alpha+\beta$.

Using $\|\mathbf{A}\|_F^2 = \operatorname{tr}(\mathbf{A}\mathbf{A}^\top)$, we expand \eqref{eq:app_obj_compact} in trace form:
\begin{equation}
\begin{aligned}
J(\mathbf{W})
&=
\operatorname{tr}\!\left[
(\mathbf{W}\mathbf{X}-\mathbf{Y})(\mathbf{W}\mathbf{X}-\mathbf{Y})^\top
\right] \\
&\quad
+(\alpha+\beta)\,
\operatorname{tr}\!\left[
(\mathbf{W}\mathbf{Z})(\mathbf{W}\mathbf{Z})^\top
\right]. 
\label{eq:app_trace}
\end{aligned}
\end{equation}

Taking the derivative with respect to $\mathbf{W}$ and using the standard rule:
\begin{equation}
\nabla_{\mathbf{W}}\operatorname{tr}(\mathbf{W}\mathbf{A}\mathbf{W}^\top\mathbf{B})
= 2\,\mathbf{B}\mathbf{W}\mathbf{A},
\end{equation}
we obtain the gradient:
\begin{equation}
\nabla_{\mathbf{W}} J
=
2(\mathbf{W}\mathbf{X}-\mathbf{Y})\mathbf{X}^\top
+ 2(\alpha+\beta)\,\mathbf{W}\mathbf{Z}\mathbf{Z}^\top.
\label{eq:app_grad}
\end{equation}
Setting $\nabla_{\mathbf{W}} J = \mathbf{0}$ yields the stationarity condition:
\begin{equation}
(\mathbf{W}\mathbf{X}-\mathbf{Y})\mathbf{X}^\top
+ (\alpha+\beta)\,\mathbf{W}\mathbf{Z}\mathbf{Z}^\top
= \mathbf{0},
\end{equation}
which can be rearranged into the linear matrix equation:
\begin{equation}
\mathbf{W}
\big(
\mathbf{X}\mathbf{X}^\top
+ (\alpha+\beta)\,\mathbf{Z}\mathbf{Z}^\top
\big)
=
\mathbf{Y}\mathbf{X}^\top.
\label{eq:app_normal_eq}
\end{equation}

The matrix in parentheses may be rank-deficient because $\mathbf{Z}=\mathbf{P}$ is a projection.  
We therefore adopt the Moore–Penrose pseudoinverse and obtain the minimum-norm solution:
\begin{equation}
\mathbf{W}^\star
=
\mathbf{Y}\mathbf{X}^\top
\big(
\mathbf{X}\mathbf{X}^\top
+ (\alpha+\beta)\,\mathbf{Z}\mathbf{Z}^\top
\big)^{+}.
\label{eq:app_W_star}
\end{equation}

Substituting back $\mathbf{X}=\mathbf{P}\mathbf{H}_m$, $\mathbf{Z}=\mathbf{P}$, $\mathbf{Y}=\mathbf{R}+\beta\mathbf{V}$, and $\mathbf{W}=\tilde{\boldsymbol{\Delta}}$ into \eqref{eq:app_W_star} gives:
\begin{equation}
\tilde{\boldsymbol{\Delta}}^{\star}
=
(\mathbf{R}+\beta\mathbf{V})
\mathbf{H}_m^\top \mathbf{P}^\top
\Big(
\mathbf{P}\mathbf{H}_m\mathbf{H}_m^\top\mathbf{P}^\top
+ (\alpha+\beta)\,\mathbf{P}\mathbf{P}^\top
\Big)^{+},
\label{eq:app_Delta_star}
\end{equation}
which coincides with Eq.~\eqref{eq:solution} in the main paper.

\section{Implementation Details}\label{appx:imp_details}
We evaluate our method on three widely adopted open-source VLMs: Qwen2-VL-7B~\cite{Qwen-VL}, MiniGPT-4-13B~\cite{MiniGPT-4}, and LLaVA-v1.5-13B~\cite{llava-v1.5}. 
Unless otherwise specified, the number of malicious samples used for optimization is set to 96.  
The steering layer is fixed at $l{=}20$ for 13B models and $l{=}14$ for the 7B model. 
For inference, LLaVA-v1.5 and Qwen2-VL use a temperature of 0.2 with top-$p{=}0.9$, while MiniGPT-4 follows its official configuration with temperature 1.0 and top-$p{=}0.9$.

\section{Additional Experimental Results}\label{appx:more_results}

\begin{table*}[h]
\tabstyle{1pt}
\centering   
\caption{Transferability performance under OOD conditions. Steering vectors are obtained from Jailbreak adversarial samples generated with $\epsilon=\tfrac{16}{255}$, using the same $\alpha$ value determined on the Jailbreak validation split. Their generalization is assessed across unseen attack types, including structure-based attacks from MM-SafetyBench~\cite{MM-SafetyBench}, perturbation-based variants of PGD, and text-only attacks. The ASR is computed using the HarmBench~\cite{HarmBench} classifier.}
\begin{tabular}{c c | c c c | c c c c c c | c}
\toprule
\multicolumn{2}{c|}{\multirow{3.5}{*}{Methods}} 
  & \multicolumn{3}{c|}{Structured-based Attack} 
  & \multicolumn{6}{c|}{Perturbation-based Attack} 
  & \multicolumn{1}{c}{Text-only Attack} \\
\cmidrule(lr){3-12}
\multicolumn{2}{c|}{} 
  & \multirow{2}{*}{SD} & \multirow{2}{*}{SD\_TYPO} & \multirow{2}{*}{TYPO} 
  & \multicolumn{2}{c}{PGD~\cite{PGD}}
  & \multicolumn{2}{c}{Auto-PGD~\cite{AutoPGD}}
  & \multicolumn{2}{c|}{MI-FGSM~\cite{MI-FGSM}} 
  & \multirow{2}{*}{GCG~\cite{Advbench}} \\
\multicolumn{2}{c|}{} 
  &  &  &  
  & $\epsilon=\tfrac{16}{255}$ & $\epsilon=\tfrac{32}{255}$
  & $\epsilon=\tfrac{16}{255}$ & $\epsilon=\tfrac{32}{255}$
  & $\epsilon=\tfrac{16}{255}$ & $\epsilon=\tfrac{32}{255}$ 
  &  \\
\midrule

{\multirow{3}{*}{{MiniGPT-4}}} 
& \multicolumn{1}{|c|}{w/o defense}         & 13.75 & 43.25 & 43.75 & 70.91 & 78.18 & 74.55 & 76.36 & 78.18 & 79.09 & 58.18 \\
& \multicolumn{1}{|c|}{ASTRA} &  3.75 &  8.75 & 11.25 &  5.45 & 12.73 &  5.45 & 10.91 & 16.37 & 13.64 &  9.09 \\
& \multicolumn{1}{|c|}{NullSteer} &  2.35 &  6.28 & 11.57 &  3.28 & 9.43 &  4.65 & 9.42 & 14.21 & 12.71 &  8.24 \\
\midrule

\multirow{3}{*}{{Qwen2-VL}} 
& \multicolumn{1}{|c|}{w/o defense}         & 20.00 & 61.25 & 38.75 & 74.55 & 80.00 & 76.37 & 77.57 & 80.00 & 78.18 & 81.82 \\
& \multicolumn{1}{|c|}{ASTRA} & 11.25 & 40.00 & 33.75 & 21.82 & 14.55 & 15.76 & 15.76 & 18.18 & 18.18 & 30.91 \\
& \multicolumn{1}{|c|}{NullSteer} & 9.86 & 38.50 & 32.53 & 19.95 & 13.86 & 14.28 & 14.35 & 16.83 & 17.28 & 28.58 \\
\midrule

\multirow{3}{*}{{LLaVA-v1.5}} 
& \multicolumn{1}{|c|}{w/o defense}         & 18.75 & 55.00 & 22.50 & 69.09 & 74.55 & 80.60 & 90.30 & 87.28 & 89.09 & 92.73 \\
& \multicolumn{1}{|c|}{ASTRA} &  8.75 & 25.00 &  6.25 &  1.82 &  1.82 &  1.21 &  0.61 &  0.00 &  0.00 & 14.55 \\
& \multicolumn{1}{|c|}{NullSteer} &  7.65 & 24.00 &  5.80 &  1.74 &  1.70 &  1.00 &  0.61 &  0.00 &  0.00 & 13.50 \\
\bottomrule
\end{tabular}
\label{tab:ood}
\end{table*}

\begin{figure*}[!t]
    \label{fig:placeholder}
    \begin{subfigure}[t]{0.45\linewidth}
        \centering
        \includegraphics[width=\linewidth]{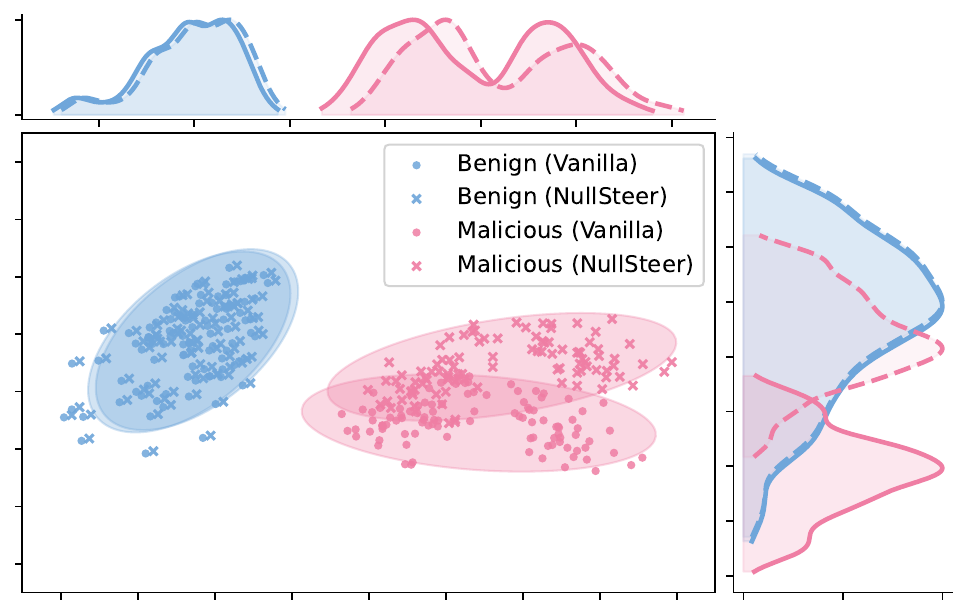}
        \caption{Activation distributions on Qwen2-VL}
        \label{fig:dis_qwen}
    \end{subfigure}
    \hfill
    \begin{subfigure}[t]{0.45\linewidth}
        \centering
        \includegraphics[width=\linewidth]{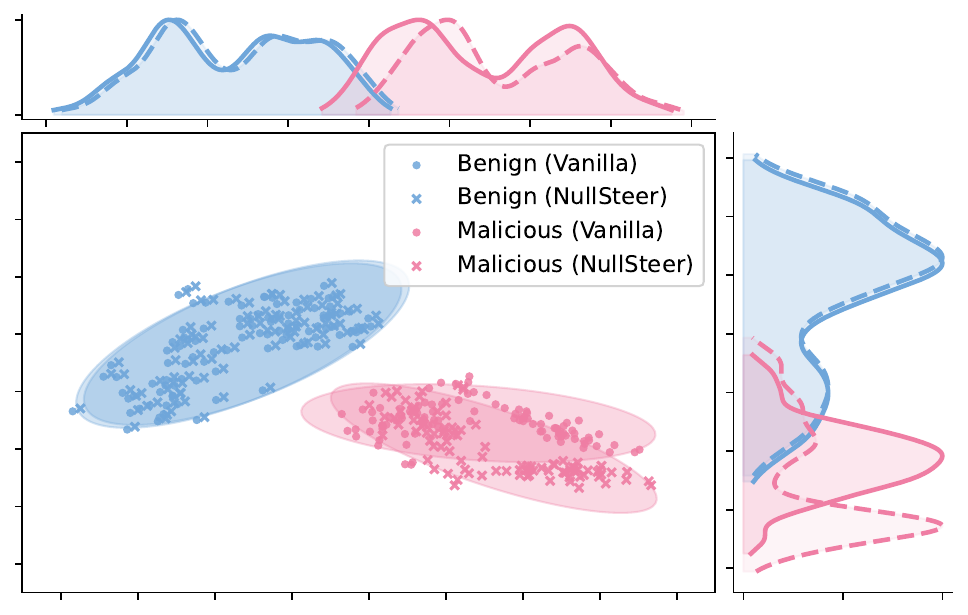}
        \caption{Activation distributions on LLaVA-v1.5.}
        \label{fig:dis_llava}
    \end{subfigure}
    \caption{Comparison of benign and malicious activation distributions across two VLMs.}
    \label{fig:dis_appx}
\end{figure*}
\noindent\textbf{OOD robustness.}
Table~\ref{tab:ood} evaluates the cross-attack generalization of steering vectors extracted from Jailbreak adversarial samples with $\epsilon=\tfrac{16}{255}$. {NullSteer} achieves consistently lower ASR across structured-, perturbation-, and text-only attacks on all three models. On MiniGPT-4, it attains 2.35\% on the SD set and 9.09\% under GCG, outperforming ASTRA (3.75\% and 8.24\%) and the undefended model (13.75\% and 58.18\%). Similar patterns are observed on Qwen2-VL and LLaVA-v1.5, where {NullSteer} notably reduces attack success under both structured and perturbation-based attacks. These results suggest that the learned steering directions capture transferable semantics of harmful behaviors rather than attack-specific perturbations.

\noindent{\textbf{Activation distribution analysis.}}
Figure~\ref{fig:dis_appx} presents the activation distributions of benign and malicious inputs on Qwen2-VL and LLaVA-1.5 before and after applying \texttt{NullSteer}.
Across both models, the benign activations under \texttt{NullSteer} closely overlap with their vanilla counterparts, forming compact clusters with nearly unchanged geometry.
This demonstrates that the null-space constraint successfully restricts the modification to directions orthogonal to benign semantics, thereby maintaining utility.
In contrast, the malicious activations exhibit a pronounced shift away from the benign region after steering, with their density moving consistently toward safer activation directions.
The clear separation between the two distributions indicates that \texttt{NullSteer} effectively isolates unsafe features while preserving the structure of benign representations, leading to stronger safety without sacrificing model behavior on harmless inputs.


\end{document}